\documentclass[11pt]{article}

\usepackage[final]{acl}

\usepackage{times}
\usepackage{latexsym}

\usepackage[T1]{fontenc}

\usepackage[utf8]{inputenc}

\usepackage{xurl}

\usepackage{microtype}
\usepackage{xspace}

\usepackage{inconsolata}

\usepackage{graphicx}
\usepackage{custom}
\title{\methodname{}: A Human-Centered Multi-Agent Writing Tutor \\for AI Research Papers on Overleaf}

\renewcommand{\thefootnote}{\fnsymbol{footnote}}
\author{%
  \textbf{Jiarui Liu\textsuperscript{1,2}, Terry Jingchen Zhang\textsuperscript{2,3}, Ryan Faulkner\textsuperscript{2}, X. Angelo Huang\textsuperscript{3,4}}\\
  \textbf{Vil\'em Zouhar\textsuperscript{4}, Dominik Glandorf\textsuperscript{5}, Isabel Dahlgren\textsuperscript{2,3,4}, Van Q. Truong\textsuperscript{2}}\\
  \textbf{Rishit Dagli\textsuperscript{2}, Yuen Chen\textsuperscript{6}, Felix Leeb\textsuperscript{7}, Punya Syon Pandey\textsuperscript{2}, Yves Bicker\textsuperscript{2,3}}\\
  \textbf{Suvajit Majumder\textsuperscript{2}, Wenyuan Jiang\textsuperscript{4}, Zeju Qiu\textsuperscript{7}, Sankalan Pal Chowdhury\textsuperscript{4}}\\
  \textbf{
  Bernhard Sch\"olkopf\textsuperscript{4,7}\footnotemark[2], Mona Diab\textsuperscript{1}\footnotemark[2], Zhijing Jin\textsuperscript{2,3,7}\footnotemark[2]}\\
  [1mm]
  \textsuperscript{1}CMU\quad
  \textsuperscript{2}Jinesis Lab, University of Toronto \& Vector Institute\quad
  \textsuperscript{3}EuroSafeAI\quad
  \textsuperscript{4}ETHZ\\
  \textsuperscript{5}EPFL\quad
  \textsuperscript{6}UIUC\quad
  \textsuperscript{7}Max Planck Institute for Intelligent Systems, T\"ubingen, Germany\\
}

\begin{document}
\maketitle
\renewcommand{\thefootnote}{\arabic{footnote}}
\setcounter{footnote}{0}
\begin{abstract}
Expert writing feedback from experienced researchers is critical for early-career scholars to improve their manuscripts, yet high-quality feedback often remains scarce because reviewing research papers is labor-intensive.
Emerging AI-powered writing assistants largely focus on grammar fixes or simulating peer review with final scores, yet they fall short of providing concrete, actionable suggestions that help students improve their papers during drafting.
We present \methodname{}, a human-centered writing assistant system that delivers actionable suggestions as Overleaf-native inline comments while leaving the actual writing entirely to human authors.
\methodname{} integrates an expert skill library carefully curated from established researchers' writing advice with 12 specialized agents covering different aspects of paper writing, such as formatting compliance, phrasing accuracy, and terminology consistency.
In a user study ($n=14$), 90.6\% of the generated comments were rated actionable and 67.5\% were rated valid, significantly outperforming a GPT-5.2 baseline uswithout the skill library.
We release \methodname{} as open source for public use.\footnote{Our code is publicly available under the AGPL-3.0 license at \url{https://github.com/jiarui-liu/overleaf}. A live demo can be accessed at \url{https://overleafmentor.ai.toronto.edu/}, and our demonstration video at \url{https://youtu.be/BD4caEJtGR0}.}

\end{abstract}

\section{Introduction}
\label{sec:introduction}

Scientific writing is a core research skill, but many junior AI researchers learn it through trial and error rather than structured mentorship.
At top NLP/AI venues such as ACL and NeurIPS, reviewers evaluate clarity, narrative, organization, and adherence to conventions alongside technical merit \citep{rogers2020reviewer, shah2022survey}. For authors without experienced mentors, weak presentation can obscure otherwise strong ideas and affect a manuscript's final acceptance \citep{widom2006tips, peyton2014write, jin2024nlp}.

Current AI writing tools do not fill this mentoring gap. Grammar assistants such as Grammarly~\citep{grammarly} and Writefull~\citep{writefull} focus mainly on sentence-level edits, while AI-powered reviewing tools~\citep{liang2024can,liu2023reviewergpt,zhou2024llm} simulate peer review and judge paper quality. Neither class of system provides drafting-stage, text-anchored feedback on narrative, organization, and technical presentation, the kind of guidance that student authors need before submission.

We introduce \methodname{}, a human-centered multi-agent writing assistant for AI scientific writing. \methodname{} delivers expert-level feedback as native inline comments on Overleaf, so authors can review suggestions within their existing collaborative workflow while retaining full control over revisions. The system combines a curated library of over 40 expert skill files with 12 specialized agents that review different aspects of a paper, including methods, results, writing style, formatting, and terminology. Each agent is guided by the relevant skills, paper type, venue expectations, and user-provided context.

We evaluate \methodname{} through a user study with 14 AI researchers who annotated comments on 80 papers from ICLR 2026 submissions and internal student drafts. Compared with direct prompting using the same LLM without the skill library, \methodname{} improves validity by 6.5 percentage points and actionability by 4.1 percentage points. We release \methodname{} as open source, providing both an Overleaf-native writing tutor and evidence that expert skill files can improve LLM feedback quality without taking revision control away from authors.

\begin{figure*}[t]
    \centering
    \includegraphics[width=0.9\linewidth]{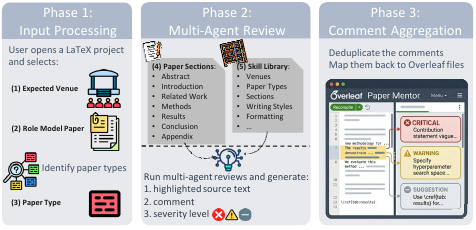}
    \caption{The three-phase pipeline of \methodname{}. In Phase 1, the system merges the uploaded LaTeX project, collects user input about the target venue and role model paper, extracts structural elements, identifies the paper type, and assigns sections to the appropriate review domains. In Phase 2, specialized review agents analyze their assigned tasks using domain-specific expertise, paper type guidelines, venue expectations, and the selected role model paper to generate structured feedback. In Phase 3, agent comments are deduplicated, consolidated, and mapped back to the original LaTeX source files for display in the Overleaf interface.}
    \label{fig:pipeline}
\end{figure*}

\section{Related Work}
\label{sec:related_work}

\noindent\textbf{LLM-Based Automated Paper Review}
Recent research has explored the use of LLMs for automated peer review, but results show that LLMs emphasize surface-level summaries over deeper methodological weaknesses and exhibit limited correlation with human scoring \citep{liang2024can, liu2023reviewergpt, zhou2024llm, yuan2022can, thakkar2025can, zhuang2025large, bougie-watanabe-2025-generative, gao2025reviewagentsbridginggaphuman, cao-etal-2025-cspaper}. AAAI-2026 introduces AI-generated supplementary reviews alongside human reviews~\citep{aaai-2026-ai-review}. Prior work has also explored multi-agent decompositions \citep{d2024marg, chamoun2024automated}, multimodal input \citep{taechoyotin2024mamorx, jin2024agentreview}, and retrieval \citep{zhu-etal-2025-deepreview}. PaperReview.ai reports near-human scoring correlation~\citep{paperreview-ai}. However, all of these systems target \textit{review-level} judgments, such as methodological soundness, novelty, and accept/reject reasoning, whereas \methodname{} generates \textit{writing-level} suggestions: concrete, text-anchored comments on writing and structure that authors need during drafting.

\noindent\textbf{Human-AI Collaborative Writing}
Commercial writing tools such as Writefull~\citep{writefull} and Grammarly~\citep{grammarly} provide grammar and vocabulary corrections. Writefull also powers Overleaf's built-in AI assistant, offering context-dependent LaTeX writing suggestions. Prism~\citep{prism} provides an alternative full LaTeX writing workspace with inline AI editing. These tools address surface-level language quality, but they do not focus on structural and organizational feedback, which is especially important in scientific writing, particularly for junior researchers. In contrast to existing commercial tools, research on human-AI writing collaboration offers design principles relevant to our work \citep{lee2024design}, showing that feedback-based assistance (commenting rather than rewriting) better preserves authorial agency \citep{dhillon2024shaping, han2024llm}. \methodname{} follows this approach by generating text-anchored comments on Overleaf, similar to the feedback a senior researcher would provide. It delivers AI domain-specific and venue-aware guidance through specialized agents informed by an expert skill library, while preserving the author's role as the person who ultimately makes the revisions.

\section{Task Definition}
\label{sec:task_definition}

Given a LaTeX project, \methodname{} generates a collection of review comments. Each comment includes four pieces of information: the source file it refers to, the character span of the highlighted text, the comment itself, and a severity label. The source file identifies which file in the project the comment belongs to. The character span marks the beginning and end positions of the highlighted passage. The comment text contains the actual feedback. The severity label indicates the importance of the issue and is one of \textsc{critical}, \textsc{warning}, or \textsc{suggestion}.

\section{System Design}
\label{sec:system_design}

\Cref{fig:pipeline} illustrates the overall architecture of \methodname{}.

\subsection{The Skill Library}
The skill library is a curated collection of expert guidance on writing strong AI research papers. It draws from two sources: internal feedback gathered from AI/ML/NLP faculty, and external publicly available writing guides by senior researchers~\citep{jin2021nlpphd,eisner2010write,peytonjones2014write,widom2006tips,wilson_scholarly,rocktaschel2022mlpaper,maddison_mlpaper,black_writing,huang2023math,acl2021ethics,boydgraber_style,parikh_shortening,forbes2021figure}.

\paragraph{Markup Taxonomy}
The sources were synthesized into a coherent skill structure by AI research experts with extensive publication experience, yielding six top-level categories: \textit{setup}, \textit{venues}, \textit{paper types}, \textit{sections}, \textit{figures and tables}, and \textit{writing style}. Topical markdown files are defined within each category according to separable sub-skills that address independent aspects of the skill topic, such as paper sections, paper types, and figure elements. Details on agent-skill assignment appear in \cref{appn:agents}.

\paragraph{Skill Authoring}
We curated a collection of publicly available writing guidance, 32 high quality published examples, and 350 reviews from 2025 conferences (NeurIPS, ICLR, and COLM) as the source material for our writing skills. We then used Claude Opus 4.5 \citep{anthropic2025claude45} to restructure and standardize this material into a consistent skill markup format. All generated markup was subsequently human reviewed and refined to ensure consistency, correctness, clarity, and conciseness throughout. The resulting library comprises over 40 skill files totaling more than 16,000 words of expert knowledge, covering paper types, target venues, individual paper sections, writing suggestions, and strategies for learning from role model papers.

\subsection{Input Processing}
The user uploads a LaTeX project and may optionally (1) select a target venue for submission and (2) provide a role model paper reflecting the style or standards they wish the system to emulate. We begin by resolving nested files to consolidate the project into a single LaTeX source file. We then extract the abstract and all sectioning headers, including sections and subsections up to the appendix. Using the merged source and the extracted structural information, the system identifies the paper type and assigns content to the appropriate review domains.

\paragraph{Paper type identification}
Different categories of AI research papers follow distinct writing conventions.
For example, a dataset paper is expected to describe data collection procedures, annotation guidelines, and evaluation details~\citep{bender2018data, gebru2021datasheets}, whereas a methods paper should clearly present the motivation, formal definition, and comparisons against relevant baselines. Our system recognizes several paper types for which we maintain dedicated expert guidance: \textit{analysis}, \textit{dataset}, \textit{method}, \textit{engineering}, \textit{interdisciplinary}, and \textit{position paper}. Using descriptions of these categories drawn from our skill library, an LLM identifies the most appropriate type. If the paper does not fit any supported category, the type is left unspecified.

\paragraph{Review domain assignment}
Following standard scientific writing conventions, we define a set of section-level review domains: abstract, introduction, related work, methods (encompassing methodology, datasets, task formulation, and preliminaries), results (encompassing experiments, findings, and discussion), conclusion (encompassing limitations, ethical considerations, and acknowledgements), and appendix. Expert guidance in the skill library is organized around these domains. Given the full draft, an LLM maps each lowest-level section header to one or more review domains. We additionally define global review domains that are not tied to any specific section, such as writing style, mathematical formatting, and table and figure captions.

\begin{table*}[t]
    \centering
    \small
    \begin{tabular}{@{}p{1.4cm}p{1.6cm}p{4cm}p{7.5cm}@{}}
        \toprule
        \textbf{Domain} & \textbf{Severity} & \textbf{Highlighted Text} & \textbf{Comment} \\
        \midrule
        Introduction & \textcolor{critical}{\textsc{critical}} & \textit{``In this paper, we propose a novel method that outperforms all baselines.''} & The contribution statement is too vague. Specify \textit{what} the method does differently and \textit{which} baselines are outperformed. A concrete summary such as ``We propose X, which achieves Y\% improvement over Z on benchmark W'' gives reviewers a clear reason to read further. \\
        \addlinespace
        Methods & \textcolor{warning}{\textsc{warning}} & \textit{``We follow standard practice for hyperparameter tuning.''} & This lacks the detail needed for reproducibility. Specify the hyperparameter search space, the selection criterion (e.g., validation loss), and the number of configurations tried. Reviewers at ACL expect full reproducibility details. \\
        \addlinespace
        Results & \textcolor{warning}{\textsc{warning}} & \textit{``Table 2 shows the results.''} & Start the results interpretation with the key finding, not a pointer to the table. For example: ``Our method achieves 42.1 F1, outperforming the strongest baseline by 3.2 points (Table~2).'' Then discuss why this improvement matters. \\
        \addlinespace
        LaTeX & \textcolor{suggestion}{\textsc{suggestion}} & \textit{``as shown in table 2''} & Use \texttt{\textbackslash cref\{tab:results\}} for consistent cross-referencing. This automatically capitalizes and formats the reference (e.g., ``Table~2'') and updates if table numbering changes. \\
        \bottomrule
    \end{tabular}
    \caption{Example comments generated by \methodname{} on a sample paper, illustrating the range of feedback across agents and severity levels. Each comment is anchored to a specific text span and provides a concrete suggestion for improvement.}
    \label{tab:example_comments}
\end{table*}

\begin{figure*}[t]
    \centering
    \includegraphics[width=0.95\linewidth]{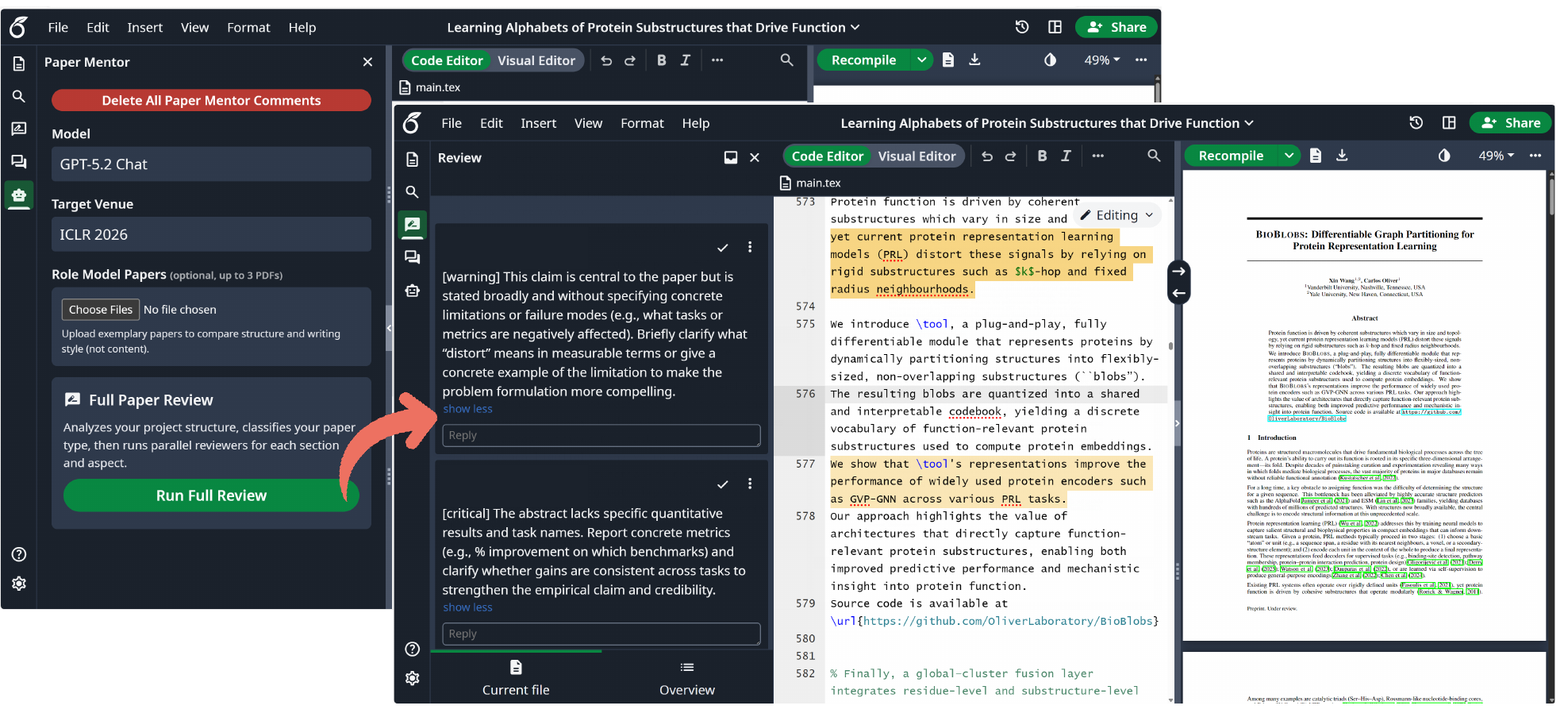}
    \caption{The \methodname{} panel within the Overleaf editor acts as a plugin that appears in the Overleaf sidebar. Left: After selecting the underlying agent model, the intended target venue for submission, and optionally one or more role model papers for reference, the user clicks ``Run Full Review'' and waits one to two minutes. Right: Once the comments are generated, the user navigates to the review panel to view all feedback produced by \methodname{}. We show this example view using the \citet{wang2025bioblobs} paper.}
    \label{fig:ui}
\end{figure*}

\subsection{Multi-Agent Review}
\label{sec:ma-review}
Because the skill library is highly modular and scientific papers are strongly structured, the review task decomposes naturally across multiple specialized agents. \methodname{} runs twelve review agents concurrently. Seven section agents each target one review domain (abstract, introduction, related work, methods, results, conclusion, and appendix); three global agents review the whole document for writing style, LaTeX and mathematical formatting, and figures and captions; and two dynamic agents are instantiated per run from the identified paper type and the selected target venue. Each agent receives the relevant LaTeX source, domain-specific skill files, paper-type-specific guidance, venue-specific expectations, and any provided role model paper; \cref{tab:agents} in \cref{appn:agents} gives the full agent skill assignment.

Skills from the library are assigned to agents according to their specialization. Section-specific agents receive only the text relevant to their assigned sections, supplemented by the abstract and introduction for context, so that their inputs remain tightly aligned with their focus. Global agents, such as those handling writing style or formatting, receive the full merged source. Each agent generates comments in the format defined in \cref{sec:task_definition}. When the input assigned to an agent exceeds a predefined length threshold, the task is further decomposed into smaller subtasks handled by lower-level sub-agents.

\subsection{Comment Aggregation}
Comments are aggregated, deduplicated, and mapped back to the original LaTeX files. Even with specialized skills, different agents may occasionally produce overlapping feedback on the same passage. We therefore remove near-duplicate comments whose highlighted spans overlap substantially and whose comment text is lexically similar. When two comments are merged, we keep the one with the higher severity, preferring section-specific agents over global agents. Finally, using the character spans produced by the agents, we map each comment back to its corresponding source file and render it in the Overleaf interface.

\section{System Demonstration}
\label{sec:demonstration}

\methodname{} is built on the open-source Overleaf Community Edition\footnote{\url{https://github.com/overleaf/overleaf}}. This choice preserves the familiar Overleaf writing environment that researchers already use, requiring no change to their existing workflow. AI-generated comments are injected via Overleaf's native ShareJS operational transformation protocol, so they appear in the review panel exactly as human reviewer comments would. \cref{tab:example_comments} presents example comments generated by different agents on a sample paper.

\paragraph{Frontend}
The frontend extends the standard Overleaf interface with a new panel in the editor's sidebar rail, implemented as a React component in TypeScript (\cref{fig:ui}). The panel exposes four controls: (1)~a model selection dropdown for the backbone LLM, (2)~an optional target venue field, (3)~an optional role model paper upload, and (4)~a ``Run Full Review'' button. Once triggered, a progress indicator is displayed until the review completes. The results appear as a collapsible summary showing the detected paper type alongside a file-by-file comment list with severity indicators. Comments are simultaneously applied to Overleaf's native review panel, where they appear with highlighted spans anchored to the corresponding locations in the LaTeX source.

\paragraph{Backend}
The backend consists of new Express.js route handlers and the review orchestration engine, implemented as ES modules within the existing Overleaf web service. When the user clicks ``Run Full Review,'' the frontend issues a POST request to the \texttt{/ai-tutor-review} endpoint carrying the project ID and selected model. The backend then retrieves all project documents, produces the merged \TeX{} file, executes the three-phase pipeline, and returns results organized by source file.

\begin{figure*}[t]
    \centering
    \includegraphics[width=\linewidth]{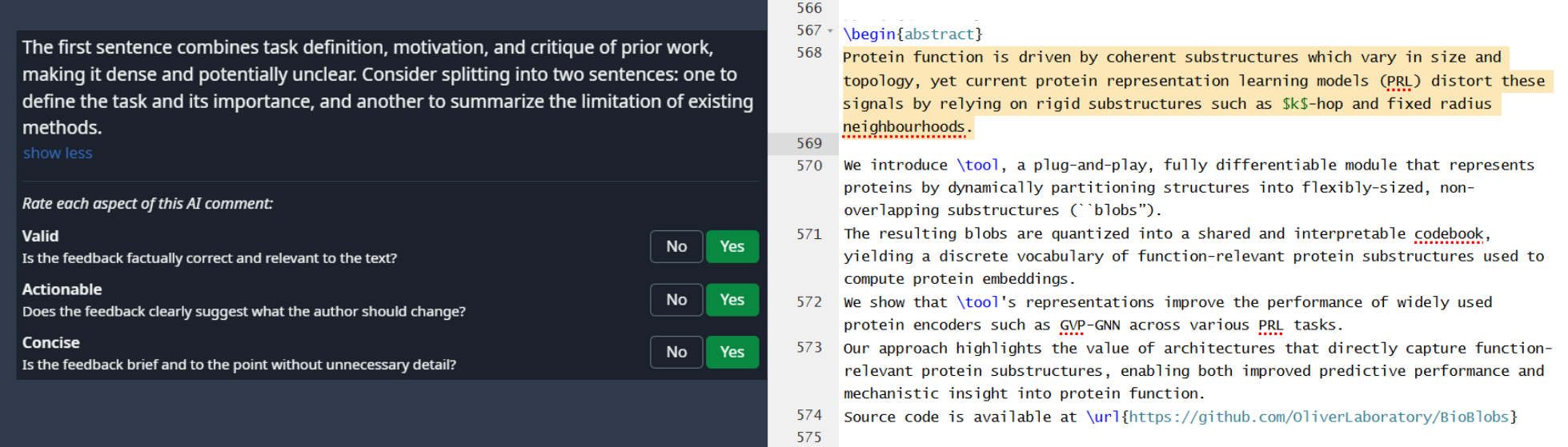}
    \caption{An example showing the annotation of a \methodname{} generated comment on our interface. The paper shown is written by \citet{wang2025bioblobs}.}
    \label{fig:annotation_example}
\end{figure*}

\section{User Study for Evaluation}
\label{sec:evaluation}

Because our core contribution lies in the expert-guided skill library, we evaluate whether our system, powered by this skill library, outperforms state-of-the-art LLM baselines in providing comments and writing suggestions.

\subsection{Experimental Setup}

\paragraph{Systems Compared}
For the baseline, we use the same LLM to directly generate comments on a paper without access to the skill library, while keeping all other prompt components identical. This ensures that any performance differences can be attributed to the incorporation of the skill library rather than other variations. We use GPT-5.2 \citep{openai2025introducinggpt52} for both \methodname{} and the baseline.

\paragraph{Dataset} We collect a total of 80 papers with compilable LaTeX sources: 10 from prior internal student submissions and 70 randomly sampled from ICLR 2026 submissions that include arXiv links with downloadable LaTeX source files. We intentionally sample from all submissions rather than only accepted papers to ensure a broad spectrum of paper quality.

\paragraph{Annotation Criteria}
We evaluate comment quality along three dimensions: \textit{validity}, \textit{actionability}, and \textit{conciseness}. \textit{Validity} asks whether the feedback is factually correct and relevant to the highlighted text. \textit{Actionability} asks whether the feedback clearly suggests what the author should change. \textit{Conciseness} asks whether the feedback is brief and to the point, without unnecessary detail or repetition.

\subsection{User Study Design}

\paragraph{Participants} We recruit 14 researchers in AI with academic backgrounds ranging from undergraduate to PhD students. Each participant logs into an assigned account on our hosted Overleaf platform and annotates four papers. For each paper, participants evaluate 60 comments: 30 generated by \methodname{} and 30 by the baseline. On the frontend, all these comments look exactly the same without layout distinctions, avoiding potential bias in the annotators' ratings.

\paragraph{Procedure} Annotators are provided with a detailed guideline document outlining the evaluation criteria. For each comment, they assess three dimensions: validity, actionability, and conciseness, selecting a binary judgment (Yes or No) for each. An example of the annotation interface is shown in \cref{fig:annotation_example}.

\paragraph{IRB} We follow the research ethics guidelines at ETH Zurich.\footnote{\url{https://ethz.ch/en/research/ethics-and-animal-welfare/research-ethics.html}} This study is exempt from ethics approval as it constitutes a survey that (1) focuses exclusively on expert knowledge (the expert acts as an informant and is not the object of the research itself); (2) offers no financial compensation; and (3) includes no experimental features such as deception, incomplete information about the study, interventions, or stimuli. We ensure (a) data protection in accordance with GDPR, (b) informed consent obtained from each expert annotator\footnote{\url{https://docs.google.com/forms/d/e/1FAIpQLSd9R7c-gltvVZz9z7njYZVs9gHGDY01Nbh0k3Jm4QGyPm8Rqg/viewform?usp=header}}, (c) strictly voluntary participation, and (d) that all collected data contain no personally identifying information.

\subsection{Results}

\begin{table}[t]
    \centering
    \resizebox{\columnwidth}{!}{%
    \begin{tabular}{lccc}
    \toprule
    \textbf{System} & \textbf{Validity} & \textbf{Actionability} & \textbf{Conciseness} \\
    \midrule
    \methodname{} & $0.675 \pm 0.023$ & $0.906 \pm 0.014$ & $0.900 \pm 0.015$ \\
    \textsc{Baseline} & $0.610 \pm 0.023$ & $0.865 \pm 0.016$ & $0.973 \pm 0.008$ \\
    \midrule
    $\Delta$ & $+0.065^{*}$ & $+0.041^{*}$ & $-0.073^{*}$ \\
    \bottomrule
    \end{tabular}
    }
    \caption{Mean human annotation ratings for \methodname{} and the baseline across three binary metrics: validity, actionability, and conciseness ($\pm$ 95\% CI). ${}^{*}p < 0.001$ (Mann–Whitney $U$ test).}
    \label{tab:annotation-ratings}
\end{table}

\cref{tab:annotation-ratings} presents the annotation results for \methodname{} and the baseline. \methodname{} significantly outperforms the direct prompting baseline in both validity and actionability. In contrast, baseline comments achieve higher conciseness on average. Overall, incorporating the skill library enables \methodname{} to generate feedback that is more accurate and more actionable.

Although the prompt provides the same instructions, incorporating the skill library increases comment length. This suggests a trade-off between conciseness and improvements in validity and actionability when adhering to structured writing guidelines.

Approximately 40\% of comments focus on the Methods and Results sections (see \cref{appn:across_domains}). When accounting for section length, \methodname{} allocates relatively more attention to high-impact sections such as the Abstract and Methods, while placing less emphasis on appendices (see \cref{appn:distribution}). Annotation scores remain consistent across major sections, indicating that comment quality generalizes well across different parts of the paper.

After completing the annotations, we qualitatively collected annotators' feedback on the comments they reviewed.\footnote{\url{https://docs.google.com/forms/d/e/1FAIpQLSe76XgkNmVhftxiV3zisP3O0f98tvDvs9TdWmbmH2d1m0vPXQ/viewform?usp=preview}} Overall, respondents viewed the AI feedback positively, with most agreeing that it mimicked a professor's tone, was easy to understand, useful for improving their paper, and generally balanced in its level of critique. The system was seen as particularly effective for clarity, depth of analysis, and grammar, and it led to moderate improvements in thesis clarity, supporting evidence, and academic rigor.

\section{Skill Library Extensibility}
\label{sec:discussion}

The skill library is designed as a living resource that can evolve over time. Researchers can extend it by contributing new skills or refining existing ones through simple text-based edits. This design makes the system easier to adapt than a fixed prompt or monolithic reviewer, since venue expectations, paper types, and disciplinary writing norms can be updated independently as the community's standards change.

We envision a community-driven development model in which writing advice from senior AI researchers across diverse subfields such as HCI, NLP, and computer vision is systematically encoded into the library. Such contributions can either enhance existing skills or be incorporated as additional skill modules. Over time, this process could turn \methodname{} from a single writing assistant into shared infrastructure for collecting, maintaining, and operationalizing practical paper-writing knowledge.

\section{Conclusion}
\label{sec:conclusion}

\methodname{} introduces a human-centered, multi-agent writing assistant that delivers expert-guided, actionable feedback directly within the Overleaf drafting workflow. By grounding specialized review agents in a curated skill library distilled from senior researchers' guidance, the system significantly improves the validity and actionability of comments over a direct prompting baseline. More broadly, our results suggest that AI writing support for research papers should move beyond generic rewriting toward structured, mentor-like feedback that helps authors revise their own work while preserving authorship and judgment.

\section*{Limitations and Future Work}
\methodname{} currently operates primarily over LaTeX source and may therefore miss issues that depend on rendered PDF output, visual figure quality, or numerical verification. Our evaluation includes 80 papers and 14 annotators, which is sufficient to demonstrate statistically significant improvements over the baseline, but does not capture the full diversity of writing styles, venues, disciplines, and researcher backgrounds. In addition, the system depends on both the coverage of the skill library and the reliability of the underlying LLM. Consequently, its feedback should be viewed as drafting assistance rather than authoritative review judgments. 

Several directions remain for future work. First, our evaluation focuses on an ablation study that isolates the contribution of the skill library by comparing \methodname{} against the same LLM without access to expert writing skills. While this design allows us to measure the effect of the skill library, it does not directly compare system-generated feedback against comments written by experienced researchers. Collecting and benchmarking against expert authored Overleaf comments would provide a stronger reference point for evaluating the overall quality and usefulness of the system.

Second, our results reveal a tradeoff between specialization and global document awareness. To improve efficiency, section-specific agents operate on limited portions of the manuscript rather than the entire paper. As a result, some validity errors occur when agents identify terms, definitions, or experimental details as missing even though they are introduced elsewhere in the document. Providing every agent with the full paper could mitigate these errors but would substantially increase computational cost and API usage. A promising direction is therefore to develop lightweight mechanisms for document-wide grounding, such as shared summaries, global definitions, or structured representations of paper content that can be efficiently accessed by all review agents.

We are actively improving \methodname{} to address these limitations and enhance the quality of its feedback. We also welcome community contributions to extend the skill library and refine the system over time, enabling it to evolve alongside the writing practices and standards of the AI research community.

\section*{Ethical Considerations}

All external papers used in our evaluation are publicly available preprints sourced from arXiv, downloaded solely for non-commercial research purposes in accordance with their respective licenses. Our user study follows the research ethics guidelines at ETH Zurich and is exempt from formal ethics review. All collected annotation data were stored securely and used exclusively for the evaluation reported in this paper.

Beyond study design, we acknowledge broader ethical considerations in deploying AI-powered writing assistance. \methodname{} is intended to support junior researchers who lack access to experienced mentors, with the goal of reducing inequalities in scientific writing guidance across institutions and geographic regions. However, we caution that over-reliance on AI feedback could inadvertently homogenize writing styles or suppress diverse rhetorical voices in scientific communication. The system generates suggestions rather than rewrites, deliberately preserving authorial agency. We also recognize that the skill library, though distilled from expert guidance, reflects the norms and conventions of predominantly English-language, Western AI venues, and may not generalize equitably to researchers writing from different cultural or disciplinary backgrounds. We encourage ongoing community contributions to the skill library to mitigate these biases over time.

\section*{Acknowledgments}
This material is based in part upon work supported by the German Federal Ministry of Education and Research (BMBF): Tübingen AI Center, FKZ: 01IS18039B; by the Machine Learning Cluster of Excellence, EXC number 2064/1 – Project number 390727645; 
and by the Canadian AI Safety Institute Research Program at CIFAR.


\bibliography{custom}

@article{shah2022survey,
    title        = {Challenges, Experiments, and Computational Solutions in Peer Review},
    author       = {Shah, Nihar B.},
    year         = 2022,
    journal      = {Communications of the {ACM}},
    volume       = 65,
    number       = 6,
    pages        = {76--87},
    doi          = {10.1145/3528086},
    url          = {https://dl.acm.org/doi/10.1145/3528086}
}

@inproceedings{cao-etal-2025-cspaper,
    title = "{CSP}aper Review: Fast, Rubric-Faithful Conference Feedback",
    author = "Cao, Lele  and
      You, Lei  and
      Team, R{\&}d",
    editor = "Flek, Lucie  and
      Narayan, Shashi  and
      Phương, L{\^e} Hồng  and
      Pei, Jiahuan",
    booktitle = "Proceedings of the 18th International Natural Language Generation Conference: System Demonstrations",
    month = oct,
    year = "2025",
    address = "Hanoi, Vietnam",
    publisher = "Association for Computational Linguistics",
    url = "https://aclanthology.org/2025.inlg-demos.2/",
    pages = "3--7"
}

@inproceedings{rogers2020reviewer,
    title        = {What Can We Do to Improve Peer Review in {NLP}?},
    author       = {Rogers, Anna and Augenstein, Isabelle},
    year         = 2020,
    booktitle    = {Findings of the Association for Computational Linguistics: {EMNLP} 2020},
    publisher    = {Association for Computational Linguistics},
    pages        = {1256--1262},
    doi          = {10.18653/v1/2020.findings-emnlp.112},
    url          = {https://aclanthology.org/2020.findings-emnlp.112/}
}

@misc{peyton2014write,
    title        = {How to Write a Great Research Paper},
    author       = {Peyton Jones, Simon},
    year         = 2014,
    howpublished = {\url{https://www.microsoft.com/en-us/research/video/how-to-write-a-great-research-paper-3/}},
    url          = {https://www.microsoft.com/en-us/research/video/how-to-write-a-great-research-paper-3/},
    note         = {Microsoft Research}
}

@misc{jin2024nlp,
    title        = {NLP PhD Global Equality: Writing Suggestions from Various Professors},
    author       = {Jin, Zhijing},
    year         = 2024,
    howpublished = {\url{https://github.com/zhijing-jin/nlp-phd-global-equality}},
    url          = {https://github.com/zhijing-jin/nlp-phd-global-equality},
    note         = {Max Planck Institute for Intelligent Systems}
}

@inproceedings{bender2018data,
    title        = {Data Statements for Natural Language Processing: Toward Mitigating System Bias and Enabling Better Science},
    author       = {Bender, Emily M. and Friedman, Batya},
    year         = 2018,
    booktitle    = {Transactions of the Association for Computational Linguistics},
    volume       = 6,
    pages        = {587--604},
    doi          = {10.1162/tacl_a_00041},
    url          = {https://aclanthology.org/Q18-1041/}
}

@article{gebru2021datasheets,
    title        = {Datasheets for Datasets},
    author       = {Gebru, Timnit and Morgenstern, Jamie and Vecchione, Briana and Vaughan, Jennifer Wortman and Wallach, Hanna and III, Hal Daum{\'e} and Crawford, Kate},
    year         = 2021,
    journal      = {Communications of the ACM},
    volume       = 64,
    number       = 12,
    pages        = {86--92},
    doi          = {10.1145/3458723},
    url          = {https://dl.acm.org/doi/10.1145/3458723}
}

@article{liang2024can,
  title={Can large language models provide useful feedback on research papers? A large-scale empirical analysis},
  author={Liang, Weixin and Zhang, Yuhui and Cao, Hancheng and Wang, Binglu and Ding, Daisy Yi and Yang, Xinyu and Vodrahalli, Kailas and He, Siyu and Smith, Daniel Scott and Yin, Yian and others},
  journal={NEJM AI},
  volume={1},
  number={8},
  pages={AIoa2400196},
  year={2024},
  publisher={Massachusetts Medical Society},
  doi={10.1056/AIoa2400196},
  url={https://doi.org/10.1056/AIoa2400196}
}

@article{liu2023reviewergpt,
  title={Reviewergpt? an exploratory study on using large language models for paper reviewing},
  author={Liu, Ryan and Shah, Nihar B},
  journal={arXiv preprint arXiv:2306.00622},
  year={2023},
  url={https://arxiv.org/abs/2306.00622}
}

@inproceedings{zhou2024llm,
  title={Is LLM a reliable reviewer? A comprehensive evaluation of LLM on automatic paper reviewing tasks},
  author={Zhou, Ruiyang and Chen, Lu and Yu, Kai},
  booktitle={Proceedings of the 2024 joint international conference on computational linguistics, language resources and evaluation (LREC-COLING 2024)},
  pages={9340--9351},
  year={2024},
  url={https://aclanthology.org/2024.lrec-main.816/}
}

@article{yuan2022can,
  title={Can we automate scientific reviewing?},
  author={Yuan, Weizhe and Liu, Pengfei and Neubig, Graham},
  journal={Journal of Artificial Intelligence Research},
  volume={75},
  pages={171--212},
  year={2022},
  doi={10.1613/jair.1.12862},
  url={https://doi.org/10.1613/jair.1.12862}
}

@article{thakkar2025can,
  title={Can LLM feedback enhance review quality? A randomized study of 20k reviews at ICLR 2025},
  author={Thakkar, Nitya and Yuksekgonul, Mert and Silberg, Jake and Garg, Animesh and Peng, Nanyun and Sha, Fei and Yu, Rose and Vondrick, Carl and Zou, James},
  journal={arXiv preprint arXiv:2504.09737},
  year={2025},
  url={https://arxiv.org/abs/2504.09737}
}

@article{zhuang2025large,
  title={Large language models for automated scholarly paper review: A survey},
  author={Zhuang, Zhenzhen and Chen, Jiandong and Xu, Hongfeng and Jiang, Yuwen and Lin, Jialiang},
  journal={Information Fusion},
  volume={124},
  pages={103332},
  year={2025},
  publisher={Elsevier},
  doi={10.1016/j.inffus.2025.103332},
  url={https://doi.org/10.1016/j.inffus.2025.103332}
}

@article{d2024marg,
  title={Marg: Multi-agent review generation for scientific papers},
  author={D'Arcy, Mike and Hope, Tom and Birnbaum, Larry and Downey, Doug},
  journal={arXiv preprint arXiv:2401.04259},
  year={2024},
  url={https://arxiv.org/abs/2401.04259}
}

@inproceedings{chamoun2024automated,
  title={Automated focused feedback generation for scientific writing assistance},
  author={Chamoun, Eric and Schlichtkrull, Michael and Vlachos, Andreas},
  booktitle={Findings of the Association for Computational Linguistics: ACL 2024},
  pages={9742--9763},
  year={2024},
  doi={10.18653/v1/2024.findings-acl.580},
  url={https://aclanthology.org/2024.findings-acl.580/}
}

@inproceedings{taechoyotin2024mamorx,
  title={MAMORX: Multi-agent multi-modal scientific review generation with external knowledge},
  author={Taechoyotin, Pawin and Wang, Guanchao and Zeng, Tong and Sides, Bradley and Acuna, Daniel},
  booktitle={Neurips 2024 Workshop Foundation Models for Science: Progress, Opportunities, and Challenges},
  year={2024},
  url={https://openreview.net/forum?id=frvkE8rCfX}
}

@inproceedings{jin2024agentreview,
  title={Agentreview: Exploring peer review dynamics with llm agents},
  author={Jin, Yiqiao and Zhao, Qinlin and Wang, Yiyang and Chen, Hao and Zhu, Kaijie and Xiao, Yijia and Wang, Jindong},
  booktitle={Proceedings of the 2024 Conference on Empirical Methods in Natural Language Processing},
  pages={1208--1226},
  year={2024},
  doi={10.18653/v1/2024.emnlp-main.70},
  url={https://aclanthology.org/2024.emnlp-main.70/}
}

@misc{gao2025reviewagentsbridginggaphuman,
      title={ReviewAgents: Bridging the Gap Between Human and AI-Generated Paper Reviews}, 
      author={Xian Gao and Jiacheng Ruan and Zongyun Zhang and Jingsheng Gao and Ting Liu and Yuzhuo Fu},
      year={2025},
      eprint={2503.08506},
      archivePrefix={arXiv},
      primaryClass={cs.CL},
      url={https://arxiv.org/abs/2503.08506}, 
}

@inproceedings{bougie-watanabe-2025-generative,
    title = "Generative Reviewer Agents: Scalable Simulacra of Peer Review",
    author = "Bougie, Nicolas  and
      Watanabe, Narimawa",
    editor = "Potdar, Saloni  and
      Rojas-Barahona, Lina  and
      Montella, Sebastien",
    booktitle = "Proceedings of the 2025 Conference on Empirical Methods in Natural Language Processing: Industry Track",
    month = nov,
    year = "2025",
    address = "Suzhou (China)",
    publisher = "Association for Computational Linguistics",
    url = "https://aclanthology.org/2025.emnlp-industry.8/",
    doi = "10.18653/v1/2025.emnlp-industry.8",
    pages = "98--116",
    ISBN = "979-8-89176-333-3"
}

@inproceedings{zhu-etal-2025-deepreview,
    title = "{D}eep{R}eview: Improving {LLM}-based Paper Review with Human-like Deep Thinking Process",
    author = "Zhu, Minjun  and
      Weng, Yixuan  and
      Yang, Linyi  and
      Zhang, Yue",
    editor = "Che, Wanxiang  and
      Nabende, Joyce  and
      Shutova, Ekaterina  and
      Pilehvar, Mohammad Taher",
    booktitle = "Proceedings of the 63rd Annual Meeting of the Association for Computational Linguistics (Volume 1: Long Papers)",
    month = jul,
    year = "2025",
    address = "Vienna, Austria",
    publisher = "Association for Computational Linguistics",
    url = "https://aclanthology.org/2025.acl-long.1420/",
    doi = "10.18653/v1/2025.acl-long.1420",
    pages = "29330--29355",
    ISBN = "979-8-89176-251-0"
}

@inproceedings{lee2024design,
  title={A design space for intelligent and interactive writing assistants},
  author={Lee, Mina and Gero, Katy Ilonka and Chung, John Joon Young and Shum, Simon Buckingham and Raheja, Vipul and Shen, Hua and Venugopalan, Subhashini and Wambsganss, Thiemo and Zhou, David and Alghamdi, Emad A and others},
  booktitle={Proceedings of the 2024 CHI Conference on Human Factors in Computing Systems},
  pages={1--35},
  year={2024},
  doi={10.1145/3613904.3642697},
  url={https://dl.acm.org/doi/10.1145/3613904.3642697}
}

@inproceedings{dhillon2024shaping,
  title={Shaping human-AI collaboration: Varied scaffolding levels in co-writing with language models},
  author={Dhillon, Paramveer S and Molaei, Somayeh and Li, Jiaqi and Golub, Maximilian and Zheng, Shaochun and Robert, Lionel Peter},
  booktitle={Proceedings of the 2024 CHI conference on human factors in computing systems},
  pages={1--18},
  year={2024},
  doi={10.1145/3613904.3642134},
  url={https://dl.acm.org/doi/10.1145/3613904.3642134}
}

@inproceedings{han2024llm,
  title     = {{LLM}-as-a-tutor in {EFL} writing education: Focusing on evaluation of student-LLM interaction},
  author    = {Han, Jieun and Yoo, Haneul and Myung, Junho and Kim, Minsun and Lim, Hyunseung and Kim, Yoonsu and Lee, Tak Yeon and Hong, Hwajung and Kim, Juho and Ahn, So-Yeon and others},
  booktitle = {Proceedings of the 1st Workshop on Customizable NLP: Progress and Challenges in Customizing NLP for a Domain, Application, Group, or Individual (CustomNLP4U)},
  pages     = {284--293},
  year      = {2024},
  url       = {https://aclanthology.org/2024.customnlp4u-1.21/}
}

@misc{writefull,
  title        = {Writefull},
  author       = {{Writefull}},
  howpublished = {\url{https://writefull.com/}},
  url          = {https://writefull.com/},
  year         = {2026},
  note         = {Accessed 2026-02-27}
}

@misc{grammarly,
  title        = {Grammarly},
  author       = {{Grammarly}},
  howpublished = {\url{https://www.grammarly.com/}},
  url          = {https://www.grammarly.com/},
  year         = {2026},
  note         = {Accessed 2026-02-27}
}

@misc{prism,
  title        = {Prism},
  author       = {{OpenAI}},
  howpublished = {\url{https://prism.openai.com/}},
  url          = {https://prism.openai.com/},
  year         = {2026},
  note         = {Accessed 2026-02-27}
}

@misc{aaai-2026-ai-review,
  title        = {{AAAI} launches {AI}-powered peer review assessment system},
  author       = {{AAAI}},
  howpublished = {\url{https://aaai.org/aaai-launches-ai-powered-peer-review-assessment-system/}},
  url          = {https://aaai.org/aaai-launches-ai-powered-peer-review-assessment-system/},
  year         = {2026},
  note         = {Accessed 2026-02-27}
}

@misc{paperreview-ai,
  title        = {PaperReview.ai},
  author       = {{Stanford}},
  howpublished = {\url{https://paperreview.ai/}},
  url          = {https://paperreview.ai/},
  year         = {2026},
  note         = {Accessed 2026-02-27}
}

@misc{jin2021nlpphd,
  author       = {Zhijing Jin},
  title        = {Resources to Help Global Equality for {PhDs} in {NLP} / {AI}},
  year         = {2021},
  howpublished = {GitHub repository},
  url          = {https://github.com/zhijing-jin/nlp-phd-global-equality},
  note         = {Open resources and information for people to succeed in PhD in CS and career in AI/NLP, including writing suggestions from various professors}
}

@misc{eisner2010write,
  author       = {Jason Eisner},
  title        = {How to Write a Paper?},
  year         = {2010},
  howpublished = {Online guide},
  url          = {https://www.cs.jhu.edu/~jason/advice/write-the-paper-first.html},
  note         = {Johns Hopkins University}
}

@misc{peytonjones2014write,
  author       = {Simon {Peyton Jones}},
  title        = {How to Write a Great Research Paper: Seven Simple Suggestions},
  year         = {2014},
  howpublished = {Slides and talk},
  url          = {https://www.cis.upenn.edu/~sweirich/icfp-plmw15/slides/peyton-jones.pdf},
  note         = {Microsoft Research. Talk available at \url{https://www.microsoft.com/en-us/research/video/how-to-write-a-great-research-paper-3/}}
}

@misc{widom2006tips,
  author       = {Jennifer Widom},
  title        = {Tips for Writing Technical Papers},
  year         = {2006},
  howpublished = {Online guide},
  url          = {https://cs.stanford.edu/people/widom/paper-writing.html},
  note         = {Stanford University}
}

@misc{wilson_scholarly,
  author       = {Shomir Wilson},
  title        = {Guide for Scholarly Writing},
  url          = {https://shomir.net/scholarly_writing.html},
  note         = {Penn State University}
}

@misc{rocktaschel2022mlpaper,
  author       = {Tim Rockt{\"{a}}schel and Jakob Foerster},
  title        = {How to {ML} Paper},
  year         = {2022},
  howpublished = {Twitter post},
  url          = {https://twitter.com/j_foerst/status/1526593779502829569},
  note         = {UCL/DeepMind and University of Oxford}
}

@misc{maddison_mlpaper,
  author       = {Chris Maddison},
  title        = {How To Write An {ML} Paper},
  howpublished = {Notion page},
  url          = {https://riemannian-connection.notion.site/How-To-Write-An-ML-Paper-1130eb91275c80c89d83d0def40f336a},
  note         = {Step-by-step writing guide}
}

@misc{black_writing,
  author       = {Michael Black},
  title        = {Writing is Laying Out Your Logical Thoughts},
  howpublished = {Twitter thread},
  url          = {https://twitter.com/Michael_J_Black/status/1598957619301187584},
  note         = {Max Planck Institute Tuebingen}
}

@misc{huang2023math,
  author       = {Jia-Bin Huang},
  title        = {How to Write Math in a Paper?},
  year         = {2023},
  howpublished = {Twitter post},
  url          = {https://twitter.com/jbhuang0604/status/1643118681960923137},
  note         = {University of Maryland}
}

@misc{acl2021ethics,
  author       = {ACL},
  title        = {Ethics {FAQ}: How to Write Ethical Considerations},
  year         = {2021},
  howpublished = {Online guide},
  url          = {https://2021.aclweb.org/ethics/Ethics-FAQ/}
}

@misc{boydgraber_style,
  author       = {Jordan Boyd-Graber},
  title        = {Style},
  howpublished = {Online guide},
  url          = {http://users.umiacs.umd.edu/~ying/static/style.html},
  note         = {University of Maryland}
}

@misc{parikh_shortening,
  author       = {Devi Parikh},
  title        = {Shortening Papers to Fit Page Limits},
  howpublished = {Medium blog post},
  url          = {https://deviparikh.medium.com/shortening-papers-to-fit-page-limits-97601318681d}
}

@misc{forbes2021figure,
  author       = {Maxwell Forbes},
  title        = {Figure Creation Tutorial: Making a Figure 1},
  year         = {2021},
  howpublished = {Online guide},
  url          = {https://maxwellforbes.com/posts/figure-creation-tutorial-making-a-figure-1},
  note         = {University of Washington}
}

@misc{openai2025introducinggpt52,
  author       = {{OpenAI}},
  title        = {Introducing GPT-5.2},
  year         = {2025},
  month        = dec,
  day          = {11},
  url          = {https://openai.com/index/introducing-gpt-5-2/}
}

@article{wang2025bioblobs,
  title={BioBlobs: Unsupervised Discovery of Functional Substructures for Protein Function Prediction},
  author={Wang, Xin and Shi, Kaiwen and Oliver, Carlos},
  journal={arXiv preprint arXiv:2510.01632},
  year={2025},
  doi={10.48550/arXiv.2510.01632},
  url={https://arxiv.org/abs/2510.01632}
}

@misc{anthropic2025claude45,
  title        = {Introducing Claude Opus 4.5},
  author       = {{Anthropic}},
  year         = {2025},
  howpublished = {\url{https://www.anthropic.com/news/claude-opus-4-5}}
}

\clearpage
\appendix

\section{Agent Configuration and Skill Assignment}
\label{appn:agents}

\methodname{} instantiates twelve review agents per run: ten with fixed scope and two configured dynamically. \cref{tab:agents} lists each agent, the text it reviews, and the expertise it draws from the skill library. Section agents review only their assigned section, supplemented by the abstract and introduction for context; global agents review the full merged source; and the figures agent reviews the extracted figure and table environments. The paper-type agent is grounded in the conventions of the type identified in Phase~1, and the venue agent in the requirements of the user-selected venue.

\begin{table*}[t]
  \centering
  \small
  \begin{tabular}{@{}p{2.7cm}p{1.5cm}p{3.2cm}p{6.6cm}@{}}
    \toprule
    \textbf{Agent} & \textbf{Class} & \textbf{Review Scope} & \textbf{Representative Skill Focus} \\
    \midrule
    Abstract & Section & Abstract & Abstract structure, specificity, and clarity \\
    Introduction & Section & Introduction & Narrative framing, motivation, and contribution statements \\
    Related Work & Section & Related work & Coverage, comparison and contrast, and citation conventions \\
    Methods & Section & Methods, task formulation, preliminaries & Methodological clarity, task formulation, and mathematical notation \\
    Results & Section & Experiments, findings, discussion & Results presentation and evidence support for claims \\
    Conclusion & Section & Conclusion, limitations, ethics & Concise conclusions, limitations, and ethical considerations \\
    Appendix & Section & Appendix & Organization of supplementary material \\
    \addlinespace
    Writing Style & Global & Full document & Grammar, tone, formality, and terminology consistency \\
    LaTeX \& Formatting & Global & Full document & LaTeX conventions, cross-referencing, and table and equation formatting \\
    Captions & Global & Figure and table environments & Caption quality \\
    \addlinespace
    Paper Type & Dynamic & Full document & Writing conventions for the identified paper type \\
    Venue & Dynamic & Full document & Formatting and content expectations of the target venue \\
    \bottomrule
  \end{tabular}
  \caption{The twelve review agents in \methodname{}, their review scope, and a representative summary of the writing expertise each draws from the skill library. Ten agents have fixed scope; the two dynamic agents are configured at runtime from the identified paper type and selected target venue.}
  \label{tab:agents}
\end{table*}

\section{Distribution of Comments and Annotation Scores Across Review Domains}
\label{appn:across_domains}

\begin{figure}[ht]
    \centering
    \includegraphics[width=\linewidth]{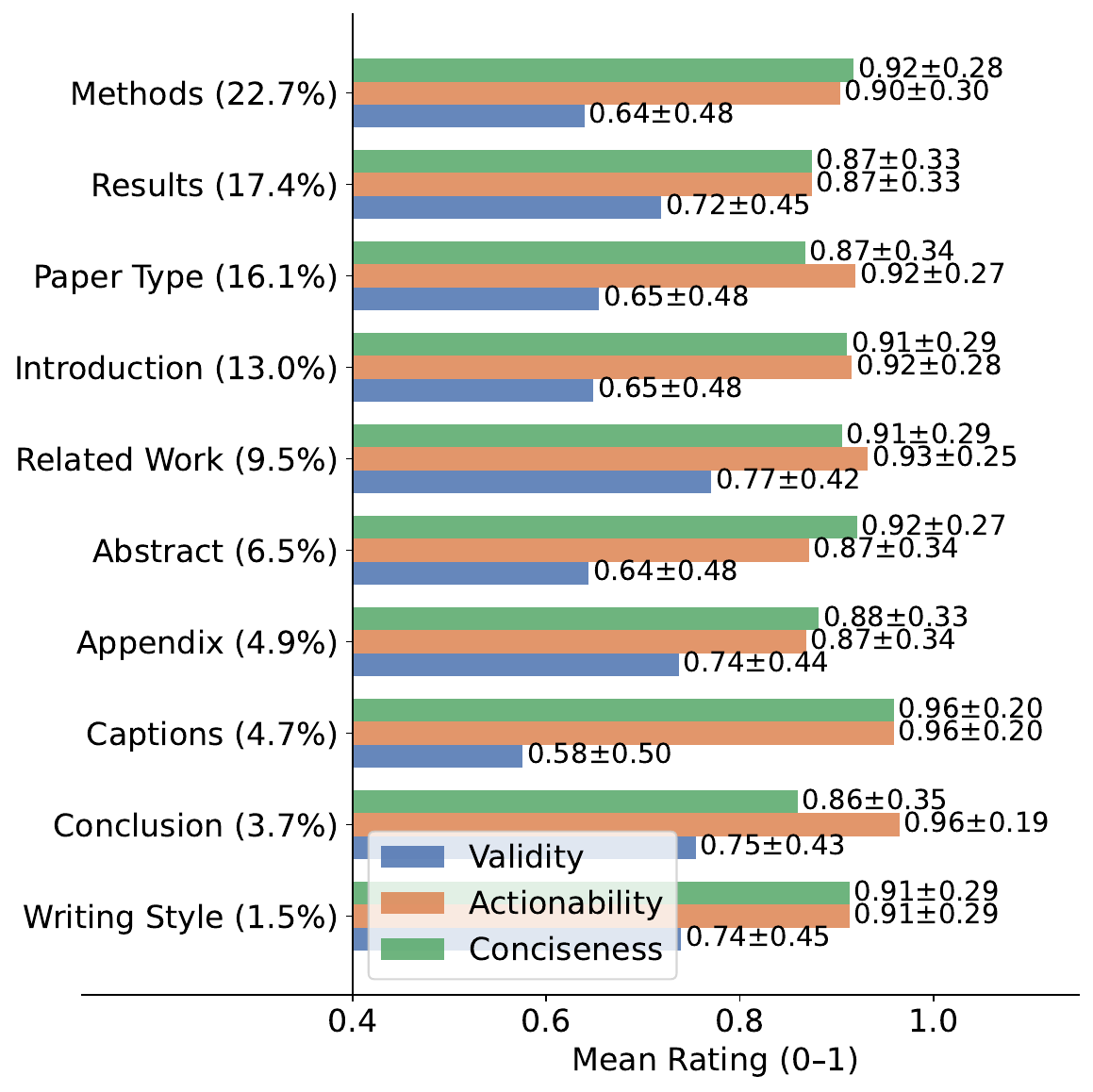}
    \caption{Distribution of generated comments across review domains, along with the mean human annotation scores for each domain on three evaluation metrics. Error bars indicate standard deviation.}
    \label{fig:annotation_by_category}
\end{figure}

\cref{fig:annotation_by_category} presents the distribution of comments generated by \methodname{} alongside the corresponding human annotation scores across different review domains.

\section{Section Level Comment Distribution vs. Text Length Distribution}
\label{appn:distribution}

\cref{tab:mean_text_vs_comments} presents the percentage distribution of comments across the main sections, compared with the proportion of text length in each section. We observe that comments are generated more frequently in core sections of the paper, such as the Abstract and Methods. This indicates that \methodname{} prioritizes more important document content, while assigning relatively fewer comments to less critical sections such as the Appendix.

\begin{table}[t]
\centering
\begin{tabular}{lrr}
\hline
Category & Text (\%) & Comments (\%) \\
\hline
Abstract & 2.5 & 8.4 \\
Introduction & 10.5 & 17.1 \\
Related Work & 5.4 & 12.3 \\
Methods & 11.3 & 24.5 \\
Results & 16.5 & 24.4 \\
Conclusion & 3.7 & 4.8 \\
Appendix & 49.9 & 8.4 \\
\hline
\end{tabular}
\caption{Percentage of total text length and percentage of total comments for each main section.}
\label{tab:mean_text_vs_comments}
\end{table}

\end{document}